# The Value of Inferring the Internal State of Traffic Participants for Autonomous Freeway Driving

Zachary N. Sunberg, Christopher J. Ho, and Mykel J. Kochenderfer

*Abstract*—Safe interaction with human drivers is one of the primary challenges for autonomous vehicles. In order to plan driving maneuvers effectively, the vehicle's control system must infer and predict how humans will behave based on their latent internal state (e.g., intentions and aggressiveness). This research uses a simple model for human behavior with unknown parameters that make up the internal states of the traffic participants and presents a method for quantifying the value of estimating these states and planning with their uncertainty explicitly modeled. An upper performance bound is established by an omniscient Monte Carlo Tree Search (MCTS) planner that has perfect knowledge of the internal states. A baseline lower bound is established by planning with MCTS assuming that all drivers have the same internal state. MCTS variants are then used to solve a partially observable Markov decision process (POMDP) that models the internal state uncertainty to determine whether inferring the internal state offers an advantage over the baseline. Applying this method to a freeway lane changing scenario reveals that there is a significant performance gap between the upper bound and baseline. POMDP planning techniques come close to closing this gap, especially when important hidden model parameters are correlated with measurable parameters.

## I. INTRODUCTION

One of the challenges in introducing autonomous automobiles is ensuring that they interact safely with human drivers. In order to navigate complex driving scenarios, human drivers routinely predict what other drivers will do and make driving decisions based on these predictions. Autonomous vehicles typically take an overly conservative approach, which can result in physical danger, reduced efficiency, and an uncomfortable experience. In a recent study, autonomous vehicles drove over 1.2 million miles without being legally responsible for any accidents. However, the autonomous vehicles actually had a higher accident rate than average for a conventional vehicle in the United States because of accidents for which they were not legally responsible [1]. This result suggests that there is significant room for improvement in autonomous-human vehicle interaction.

One approach to improve interaction would be to program ad-hoc logic for each situation into the vehicles. However, this approach is time-consuming and error prone, and edge cases that the programmers have not foreseen can present a safety risk. Furthermore, this approach limits the performance of the system to the capability of the human programmer. In contrast, artificial intelligence and machine learning techniques have the potential to provide a more robust approach to such decision-making tasks. This paper explores techniques based on Markov decision processes (MDPs) and partially observable Markov decision processes (POMDPs) [2].

POMDPs are particularly well suited for modeling decisions for autonomous vehicles because it explicitly captures the limitations of the vehicle's sensors in measuring the relevant state variables [3]–[5]. Though sensors can accurately measure many of the relevant variables pertaining to the physical state of the vehicles, the internal state (e.g., intentions and aggressiveness) of other drivers and road users can only be indirectly inferred [4]–[7]. The hypothesis explored in this paper is that inferring and planning with an estimate of the internal states of the traffic participants will improve safety and efficiency.

Driving strategies derived from MDPs and POMDPs depend on several ingredients to be successful. First, an accurate stochastic model of the environment, including the behavior of other drivers is necessary. Though this paper uses a very simple model, there has been significant work on creating better models in recent years [8]–[10], and the POMDP planning methods that we use can easily be adapted to use these new models. POMDP methods also require a reward function. In some cases, constructing the reward function that balances different objectives (e.g., safety and efficiency) is straightforward. However, in other cases, human preferences can be difficult to quantify and encode into a reward function. For these cases, inverse reinforcement learning can be used to determine a suitable reward function based on data of how human drivers act [11], [12].

Before investing the effort required to develop and test a POMDP-based decision making system for real autonomous vehicles, it is important to quantify the potential performance improvement. This paper presents a method that involves comparing solutions obtained from several variations of Monte Carlo Tree Search [13]. For this research, we have chosen to investigate these ideas in the context of making lane changes on a freeway (a situation that has been anecdotally noted to be difficult [14]). We present a method for quantifying the performance gains that could result from perfect estimation of and planning with hidden behavior model parameters. In addition, we show that when model parameters are correlated, estimating the parameters online using physical measurements can greatly improve performance. Planning using a POMDP problem formulation that dynamically takes uncertainty into account can further improve performance.

## II. MODEL

The focus of this paper is on freeway driving. We investigate a scenario in which a vehicle must navigate from the rightmost to the leftmost lane of a four lane freeway as quickly as possible while maintaining safety and comfort.

Throughout this section, $x$ denotes position in the *longitudinal* direction, that is, the direction that the cars move along the road in meters, and $y$ denotes position in the *lateral* direction, that is, the lane the car occupies in lane units. The problem can be stated as a discrete-time POMDP defined by the tuple $(S, A, T, R, O, Z)$, which consists of

- The state space, $S$: A system state,

$$s = (q_0, \{(q_i, \theta_i)\}_{i \in 1..N}) \in S,$$

  consists of the physical state of the ego vehicle ($q_0$), and physical state and behavior model for each of the $N$ other cars in the scene. The physical state,

$$q_i = (x_i, y_i, \dot{x}_i, \dot{y}_i),$$

  consists of the car's longitudinal and lateral position and velocity. The internal state (behavior model parameters), $\theta_i$, is drawn from a set of behaviors $\Theta$.
- The action space, $A$: An action, $u = (\ddot{x}_e, \dot{y}_e) \in A$, consists of the longitudinal acceleration and lateral velocity of the ego vehicle. The action space is discrete and pruned to prevent crashes (see Section II-C).
- The state transition model, $T : S \times A \times S \to \mathbb{R}$: The value $T(s, u, s')$ is the probability of transitioning to state $s'$ given that action $u$ is taken by the ego at state $s$. This function is implicitly defined by a generative model that consists of a state transition function, $F(\cdot)$, and a stochastic noise process (see Section II-B).
- The reward model, $R : S \times S \to \mathbb{R}$: The reward function, defined in Section II-D, rewards time spent in the target left lane and penalizes strong braking.
- The observation space, $O$: An observation, $o \in O$ consists of the physical states of all of the vehicles, that is $o = \{p_i\}_{i \in 1..N}$. No information about the internal state is directly included in the observation.
- The observation model, $Z : S \times O \to \mathbb{R}$: The value $Z(s', o)$ is the probability of receiving observation $o$ when the system transitions to state $s'$. In these experiments, the physical state is assumed to be known exactly, though it is not difficult to relax this assumption.

The remainder of this section elaborates on this model.

### A. Driver Modeling

The driver models for each car have two components: an acceleration model that governs the longitudinal motion and a lane change model that determines the lateral motion. In this paper, the acceleration model is the Intelligent Driver Model (IDM) [15], and the lane change model is the "Minimizing Overall Braking Induced by Lane change" (MOBIL) model [16]. Both of these models have a small number of parameters that determine the behavior of the drivers.

*1) IDM:* The IDM Model was developed as a simple model for "microscopic" simulations of traffic flows and is able to reproduce some phenomena observed in real-world traffic flows. It determines the longitudinal acceleration for a human-driven car, $\ddot{x}$, based on the desired distance gap to the preceding car, $g$, the absolute velocity, $\dot{x}$, and the velocity relative to the preceding car $\Delta \dot{x}$. The longitudinal acceleration is governed by the following equation:

$$\ddot{x}_{\text{IDM}} = a \left[ 1 - \left( \frac{\dot{x}}{\dot{x}_0} \right)^\delta - \left( \frac{g^*(\dot{x}, \Delta \dot{x})}{g} \right)^2 \right], \quad (1)$$

where $g^*$ is the desired gap given by

$$g^*(\dot{x}, \Delta \dot{x}) = g_0 + T\dot{x} + \frac{\dot{x} \Delta \dot{x}}{2\sqrt{ab}}. \quad (2)$$

Brief descriptions and values for the parameters not defined here are provided later in Table I.

A small amount of noise is also added to the acceleration

$$\ddot{x} = \ddot{x}_{\text{IDM}} + \frac{\sigma_{\text{vel}}}{\Delta t} w, \quad (3)$$

where $w$ is zero-mean, normally distributed random variable with unit standard deviation. The $w$ for each car and time step are independent of all others. Since $w$ can take arbitrarily large values, crashes can technically occur. However, at each transition, if the additional noise is sufficiently large to cause a collision, $w$ is artificially reduced to prevent collision. In practice this occurs very rarely.

*2) MOBIL:* The MOBIL model makes the decision to change lanes based on maximizing the acceleration for the vehicle and its neighbors. When considering a lane change, MOBIL first ensures that the safety criterion $\tilde{\ddot{x}}_{\text{follow}} \geq -b_{\text{safe}}$, where $\ddot{x}_n$ will be the acceleration of the following car if the lane change is made and $b_{\text{safe}}$ is the safe braking limit. It then makes the lane change if the following condition is met

$$\tilde{\ddot{x}}_c - \ddot{x}_c + p \left( \tilde{\ddot{x}}_n - \ddot{x}_n + \tilde{\ddot{x}}_o - \ddot{x}_o \right) > \Delta a_{\text{th}} \quad (4)$$

where the quantities with tildes are calculated assuming that a lane change is made, the quantities with subscript $c$ are quantities for the car making the lane change decision, those with $n$ are for the new follower, and those with $o$ are for the old follower. The parameter $p \in [0, 1]$ is the politeness factor, which represents how much the driver values allowing other vehicles to increase their acceleration. The parameter $\Delta a_{\text{th}}$ is the threshold acceleration increase to initiate a lane changing maneuver. Parameter values are listed in Table I.

### B. Physical Dynamics

The physical dynamics are simplified for the sake of computational efficiency. Time is divided into discrete steps of length $\Delta t$. The longitudinal dynamics assume constant acceleration, and the lateral dynamics assume constant velocity over a time step, that is

$$\begin{aligned} x' &= x + \dot{x}\Delta t + \frac{1}{2}\ddot{x}\Delta t^2 \\ \dot{x}' &= \dot{x} + \ddot{x}\Delta t \\ y' &= y + \dot{y}\Delta t. \end{aligned}$$

There is a physical limit to the braking acceleration, $b_{\max}$. Lateral velocity is allowed to change instantly because cars on a freeway can achieve the lateral velocity needed for a lane change in time much shorter than $\Delta t$ by steering. If MOBIL determines that a lane change should be made, the lateral velocity, $\dot{y}$, is set to $\dot{y}_{\text{lc}}$. Lane changes are not allowed to reverse. Once a lane change has begun, $\dot{y}$ remains constant until the lane change is completed (this is the reason that $\dot{y}$ is part of the state). When a vehicle passes over the midpoint of a lane, lateral movement is immediately stopped so that lane changes always end at exactly the center of a lane.

Since MOBIL only considers cars in adjacent lanes, there must be a coordination mechanism so that two cars do not converge into the same lane simultaneously. In order to accomplish this, if two cars begin changing into the same lane simultaneously, and the front vehicle is within $g^*$ of the rear vehicle, the rear vehicle's lane change is canceled.

In order to reduce the computational demands of decision-making, only 50 m of road in front of the ego and 50 m behind are modeled. Thus, a model for vehicle entry into this section is needed. If there are fewer than $N_{\max}$ vehicles on the road, a new vehicle is generated. First, a behavior for the new vehicle is drawn from $\Theta$, and the initial speed is set to $\dot{x}_0 + \sigma_{\text{vel}} w_0$, where $\dot{x}_0$ is the desired speed from the behavior model and $w_0$ is a zero-mean, unit-variance, normally distributed random variable that is independent for each car. If this speed is greater than the ego's speed, the new vehicle will appear at the back of the road section; if it is less, it will appear at the front. For each lane, $g^*$ is calculated, either for the new vehicle if the appearance is at the back or for the nearest following vehicle if the appearance is at the front. The new vehicle appears in the lane where the clearance to the nearest car is greatest. If no clearance is greater than $g^*$, the new vehicle does not appear.

For convenience, throughout this paper, the behavior described so far will be denoted compactly by the state transition function

$$s' = F(s, u, w). \tag{5}$$

### C. Action Space for Crash-Free Driving

At each time step, the planner for the ego must choose the longitudinal and lateral acceleration. For simplicity, the vehicle chooses from up to ten discrete actions. The vehicle may make an incremental decrease or increase in speed or maintain speed, and it may begin a left or right lane change or maintain the current lane. The combination of these adjustments make up nine of the actions. The final action is a braking action determined dynamically based on the speed and position of the vehicle ahead. At each time step, the maximum permitted acceleration, $a_{\max}$, is the maximum acceleration that the ego could take such that, if the vehicle ahead immediately begins braking at the physical limit, $b_{\max}$, to a stop, the ego will still be able to stop before hitting it without exceeding physical braking limits itself. The braking action is $(\ddot{x}_e, \ddot{y}_e) = (\min\{a_{\max}, -b_{\text{nominal}}\}, 0)$.

The inclusion of the dynamic braking action guarantees that there will always be an action available to the ego to avoid a crash. At each step, the action space is pruned so that if $\ddot{x}_e > a_{\max}$ or if a lane change leads to a crash, that action is not considered. Since the IDM and MOBIL models are both crash-free [17], and actions that lead to crashes for the ego are not considered, no crashes occur in the simulation. Eliminating crashes in our model is justifiable because it is likely that in an actual autonomous vehicle a high-level planning system would be augmented with a low-level crash prevention system to increase safety and facilitate certification. In addition, it is difficult to model driver behavior in the extraordinary case of a crash.

### D. Reward Function and Objectives

The qualitative objectives in solving this problem are to reach the target lane as quickly as possible and increase the comfort and safety of both the ego and the other nearby vehicles. Thus, the following two metrics will be used to evaluate planning performance: 1) the average time taken for the ego to reach the target lane, and 2) the number of hard braking maneuvers that *any* vehicle undertakes during the time that it takes for the ego to reach the target lane. A hard braking maneuver occurs any time that $\ddot{x} < -b_{\text{hard}}$, where $b_{\text{hard}}$ is chosen to be an uncomfortably abrupt deceleration. The number of hard braking maneuvers is a proxy for both safety and comfort.

In order to encourage the planner to choose actions that will maximize these metrics, the reward function for the POMDP is defined as follows:

$$R(s, s') = \mathbf{1}(y_e = y_{\text{target}}) - \lambda \sum_{i=1}^{N} \mathbf{1}(\dot{x}'_i - \dot{x}_i < -b_{\text{hard}} \Delta t). \tag{6}$$

Thus, a reward is generated for each step in the target lane, and a cost is accrued for each braking maneuver. The weight $\lambda$ balances the competing goals.

### E. Initial Scenes

Initial scenes for the simulations are generated by beginning a simulation with only the ego on the road section. We then simulate 200 steps with the ego maintaining the current lane and using the IDM model with typical parameter values.

## III. SOLUTION APPROACHES

Monte Carlo tree search (MCTS) is one of the most widely used and effective methods for solving decision-making problems online [13]. MCTS creates a tree consisting of alternating levels of nodes corresponding to actions and states. Estimates of the value (expected discounted cumulative rewards) are maintained for each action node. In this paper, we consider four variants of MCTS to solve different versions of the problem. All of the variants make use of the upper confidence tree (UCT) [13] and double progressive widening (DPW) [18], [19] modifications to MCTS.

When building the tree, UCT expands the action nodes that maximize an upper confidence estimate

$$UCB(s, u) = \tilde{Q}(s, u) + c\sqrt{\frac{\ln N(s)}{N(s, u)}},$$

where $\tilde{Q}(s,u)$ is the estimate of the action-value function obtained through rollout simulations and tree search, $N(s,u)$ is the number of times action $u$ has been tried from state $s$, $N(s) = \sum_{u \in \mathcal{A}} N(s,u)$, and $c$ is the exploration constant. This efficiently balances exploration of the tree towards promising regions of the search space.

DPW is used to govern the growth of the tree in large state spaces. If MCTS is applied to a large state space (such as the continuous state space of this lane changing problem) without a widening control mechanism in place, too many new states will be visited from each action node. This will result in a shallow tree that does not define a good policy. A DPW tree contains state-action nodes corresponding to each action that has been tried in each state. In order to control widening, DPW limits the number of children of a state-action node $(s,u)$ to

$$k\,N(s,u)^\alpha, \qquad (7)$$

where $k$ and $\alpha$ are tunable parameters. When there are fewer children than this limit, a new state is generated by simulating the dynamics and a child corresponding to the new state is added. As $N(s,u)$ grows, so does the number of children, allowing for gradual widening of the tree. DPW also limits the number of actions that are considered from each state ("double" in DPW refers to limiting the widening at both state and action nodes), but this is less important for this problem because the action space is small.

The remainder of this section discusses three different approaches for planning with the internal states of the traffic participants.

### A. Approach 1: Static Assumed Behavior (SAB)

A performance baseline is established by planning as if all cars behave according to a single static "normal" internal state (see Table I). In this case, the problem is an MDP, which is solved using the MCTS-DPW algorithm.

### B. Approach 2: Most Likely Model Predictive Control (MLMPC)

Since information about the human's internal state can be inferred by observing the car's physical motion, performance superior to the SAB baseline can be achieved by estimating $\theta$ online. This is accomplished with a particle filter [20]. Filtering is independent for each car, but all of the behavior parameters for a given car are estimated jointly. There are two versions of the filter. In the first version, the behavior parameters are assumed to be uncorrelated, so a particle, $\hat{\theta}$ consists of values of all model parameters. In the second version, all parameters are assumed perfectly correlated (see Section IV-A), so a particle consists of only a single value, the "aggressiveness".

The belief at a given time consists of the exactly known physical state, $q$, and a collection of $M$ particles, $\{\hat{\theta}^k\}_{k=1}^{M}$ along with associated weights $\{W^k\}_{k=1}^{M}$. To update the belief when action $u$ is taken, $M$ new particles are sampled with probability proportional to the weights, and sampled noise values $\{\hat{w}^k\}_{k=1}^{M}$ are used to generate new states according to $\hat{s}^{k\prime} = F((q, \hat{\theta}^k), u, \hat{w}^k)$. The new weights are determined by approximating the conditional probability of the particle given the observation:

$$W^{k\prime} = \begin{cases} \exp\left(-\frac{(\dot{x}'-\hat{\dot{x}}')^2}{2\sigma_{\text{vel}}^2}\right) & \text{if } y' = \hat{y}' \\ \gamma_{\text{lane}} \exp\left(-\frac{(\dot{x}'-\hat{\dot{x}}')^2}{2\sigma_{\text{vel}}^2}\right) & \text{o.w.} \end{cases} \propto \Pr\left(\hat{\theta}^k \middle| o\right)$$

where $\dot{x}'$ and $y'$ are taken from the observation, $\hat{\dot{x}}'$ and $\hat{y}'$ are from $\hat{s}^{k\prime}$, the exponential expression is proportional to the Gaussian probability density function (from the acceleration noise), and $\gamma_{\text{lane}} \in [0,1]$ is a hand-tuned parameter that penalizes incorrect lane changes.

In order to prevent particle deprivation, during the resampling step, Gaussian noise with standard deviation proportional to the sample standard deviation of the current particle set is added to 10 % of the new samples.

Model predictive control (MPC) is a widely used family of control techniques that use an imperfect model and feedback measurements to choose actions [21]. At each time step, a model predictive controller calculates a sequence of control actions that will maximize a reward function of the states visited up to a future horizon given that the system behaves according to a model. The first control action in this optimized sequence is executed, and the process is repeated after a new measurement is received.

In the most likely model predictive control (MLMPC) approach, we use this particle filter to estimate the internal state for each driver. At each step, the model used for MPC is the MDP that results from assuming that each driver has the internal state corresponding to the single particle with the highest weight. Each time a new observation is received, the particle filter is updated and MCTS-DPW determines the best action for the resulting MDP.

### C. Approach 3: POMCP with Double Progressive Widening

The MCTS algorithm has been extended to handle domains with state uncertainty in the partially observable Monte Carlo planning (POMCP) algorithm [22]. The root node of a POMCP tree corresponds to the current belief about the state maintained by the particle filter described above. In place of the state nodes of MCTS, POMCP uses history nodes that correspond to the sequence of actions and observations required to reach the node from the root. At each history node, POMCP represents the current belief using a collection of unweighted particles.

In order to address the challenges with the branching factor when using MCTS with continuous state spaces, we have adapted POMCP to use DPW. Equation (7) is used to limit the number of children of each action node. In cases when a new history node is not generated, a previously generated history node is selected and the next state is sampled from the particle collection corresponding to that node. The authors are not aware of any previous research that uses POMCP with DPW, so the algorithm's general properties have not yet been thoroughly studied.

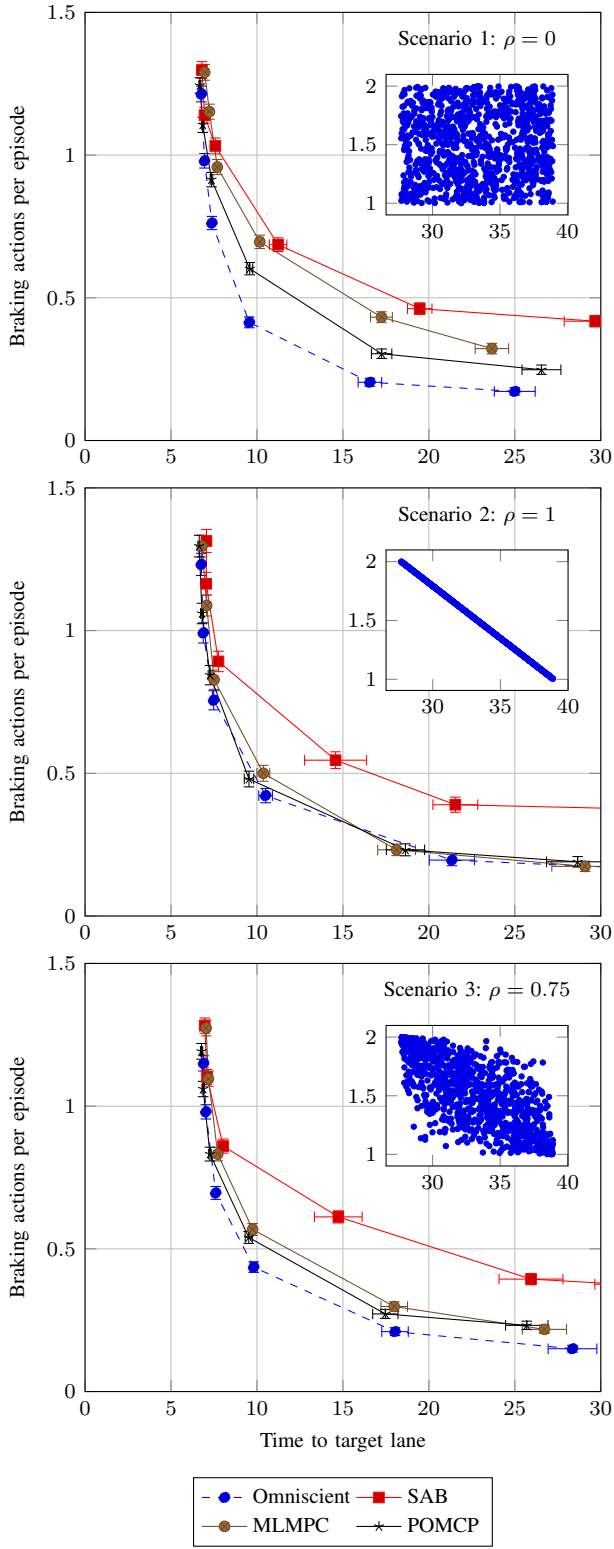

Fig. 1. Performance curves for different solvers. The inset plots illustrate the parameter correlation. The horizontal axis shows $\dot{x}_0$ in m/s and the vertical axis shows $T$ in seconds. Error bars indicate the standard error of the mean, that is $\sigma_{\text{sample}}/\sqrt{n}$, where $\sigma_{\text{sample}}$ is the sample standard deviation and $n$ is the number of simulations.

## IV. RESULTS AND DISCUSSION

The computational results from this study are designed to meet the two goals of 1) quantifying the size of the gap between the baseline control algorithm and the maximum potential lane change performance and 2) showing which cases internal state estimation and POMDP planning can approach the upper bound on performance. Experiments are carried out in three scenarios, each with a different distribution of internal states. In each of these scenarios, each of the three solution methods described in Section III are compared with an approximate upper performance bound obtained by planning with perfect knowledge of the behavior models.

### A. Driver Model Distribution Scenarios

We studied three internal state distribution scenarios. In all of these scenarios, drivers behave according to the IDM and MOBIL models presented in Section II-A, however the IDM and MOBIL parameter values are distributed differently.

Table I shows typical parameter values for aggressive, timid, and normal drivers. The values are taken from [17], but some have been adjusted slightly so that the parameters for the normal driver are exactly half way between values for the timid and aggressive drivers. In all three of the scenarios, the *marginal* distributions of the parameters are uniformly distributed between the aggressive and timid values. The difference between the scenarios is the correlation of the parameter values. In Scenario 1, all of the parameters are independently distributed. In Scenario 2, all of the parameters are perfectly correlated so that all parameters are deterministic functions of the aggressiveness of the driver. Scenario 3 uses a distribution between these two extremes. In this scenario, values are drawn from a Gaussian copula with covariance matrix with 1 along the diagonal and a correlation, $\rho \in (0, 1)$, elsewhere. The values are then scaled and translated to lie between the aggressive and normal limits. For Scenario 3, the value of $\rho$ is 0.75, and Scenarios 1 and 2 can be thought of as limiting cases where $\rho$ approaches 0 and 1, respectively. In Scenario 1, the first version of the particle filter, which estimates all of the model parameters jointly, is used, whereas in Scenarios 2 and 3, the second version of the particle filter that assumes fully correlated parameters is used, that is, it only estimates a single "aggressiveness" parameter for each car. The small inset plots in Fig. 1 illustrate the level of correlation by plotting sampled values of two of the parameters.

### B. Performance Results

Figure 1 shows the performance results from the simulations. Since the conservativeness of all of the solution techniques can be adjusted by changing $\lambda$ in (6), 500 simulations were carried out using a reward function with $\lambda$ set to the values in Table II. The average time taken to reach the target lane and the average number of braking actions per episode are plotted, and a linear interpolation between these points yields an approximation of the Pareto-optimal frontier.

TABLE I
IDM AND MOBIL PARAMETERS FOR DIFFERENT DRIVER TYPES.

| IDM Parameter | | Aggressive | Timid | Normal |
|---|---|---|---|---|
| Desired speed (m/s) | $\dot{x}_0$ | 38.9 | 27.8 | 33.3 |
| Desired time gap (s) | $T$ | 1.0 | 2.0 | 1.5 |
| Jam distance (m) | $g_0$ | 0.0 | 4.0 | 2.0 |
| Max acceleration (m/s$^2$) | $a$ | 2.0 | 0.8 | 1.4 |
| Desired deceleration (m/s$^2$) | $b$ | 3.0 | 1.0 | 2.0 |
| MOBIL Parameter | | Aggressive | Timid | Normal |
| Politeness | $p$ | 0.0 | 1.0 | 0.5 |
| Safe braking (m/s$^2$) | $b_{safe}$ | 3.0 | 1.0 | 2.0 |
| Acceleration threshold (m/s$^2$) | $a_{thr}$ | 0.0 | 0.2 | 0.1 |

TABLE II
VARIOUS SIMULATION PARAMETERS

| Parameter | Symbol | Value |
|---|---|---|
| Simulation time step | $\Delta t$ | 0.75 s |
| Max vehicles on road | $N_{max}$ | 10 |
| Lane change rate | $\dot{y}_{lc}$ | 0.67 lanes/s |
| Velocity noise standard deviation | $\sigma_{vel}$ | 0.5 m/s |
| Physical braking limit | $b_{max}$ | 8.0 m/s$^2$ |
| Penalized hard braking limit | $b_{hard}$ | 4.0 m/s$^2$ |
| UCT exploration parameter | $c$ | 5 |
| DPW linear parameter | $k$ | 4 |
| DPW exponent parameter | $\alpha$ | 0.125 |
| MCTS search depth | | 20 |
| MCTS-DPW iterations per step | | 500 |
| POMCP iterations per step | | 2500 |
| Particle filter wrong lane factor | $\gamma_{lane}$ | 0.2 |
| Number of Particles (Joint Parameter Filter) | $M$ | 1000 |
| Number of Particles (Aggressiveness Filter) | $M$ | 500 |
| Reward ratios for points on Pareto curves | $\lambda$ | 1, 2, 4, 8, 16, 32 |

In all cases, there is a significant performance gap between the upper bound and the baseline planning algorithm that assumes all cars behave according to the normal values of the driver model parameters. For example, in Scenario 1, if it is acceptable to have an average of 0.5 hard braking actions per episode, then the omniscient planner is able to reach the lane in an average of 9.0 s (50 %) less time than the baseline SAB planner. Alternatively, if the vehicle is required to complete the lane changing task in an average time of 10 s, then the average number of hard brakes per episode would be decreased by 0.40 (50 %) with perfect internal state knowledge. The performance gap between the baseline and the bound decreases slightly as the correlation, $\rho$, increases.

Perhaps the most important result that can be gleaned from this data is the relationship between the correlation and the extent to which more advanced planning methods can close the gap to the bound. When the parameters are uncorrelated, POMCP is able to reduce the gap to approximately half its size, while MLMPC struggles to perform much better than the baseline. However, in the fully correlated scenario, both of the approaches that try to estimate the internal state are able to fully close the gap. The reason for this lies primarily in the difficulty in estimating the behavior parameters. In these tests, the human-driven cars spend most of their time near their desired speed, $v_0$. This makes it relatively easy to estimate $v_0$ using the particle filter, but difficult to measure the other parameters that determine how the vehicle will react to the ego. In the fully correlated case, estimating $v_0$ also gives the exact values of all other parameters, so POMCP and the MLMPC approach are essentially planning with known models. The fully correlated case is rather unrealistic, but the results from Scenario 3 indicate that even if the parameters are only partially correlated, most of the losses caused by not knowing the model can be recovered.

The gap between the POMCP and most likely behavior approaches also indicates the relative importance of the two tasks of estimation and planning when accounting for internal state uncertainty. Since the POMCP approach plans with the full distribution of possible internal states, it takes uncertainty and future feedback into account when planning. The MLMPC approach plans as if the behavior models are exactly known and does not incorporate possible future feedback. It could be called an "open loop" planner. Of course, when the model is easy to estimate with high certainty in the fully correlated scenario, there is little difference between the POMCP and MLMPC performance. However, in Scenario 1, where there is large uncertainty about the model, POMCP does show a significant improvement.

The Julia source code for the experiments can be found at https://github.com/sisl/Multilane.jl, and the solver software is part of the POMDPs.jl package (https://github.com/JuliaPOMDP/POMDPs.jl).

V. CONCLUSION

This paper has presented a method for quantifying the performance that could be gained by a planner omniscient to the internal state of other vehicles compared to a baseline. In the simplified case that we investigated, this gap proved to be significant. Moreover, we showed that the use of a simple particle filter can offer greatly improved performance, especially when internal state features are highly correlated, and planning using a POMDP model can further reduce the gap to the bound.

The method described in this paper is extremely flexible and adaptable. One only needs to define a generative model for states, rewards, and observations and possibly tune some algorithm parameters to adapt it to a new domain. This makes it useful as a preliminary analysis tool for deciding which control approach to implement on an actual vehicle.

For the particular case of freeway lane changing, the most important result from this research is that the distribution of behavior models in the driver population is a critical factor in determining the performance improvement that can be gained by estimating the internal state online and planning in a way that takes uncertainty in the model into account. If internal states that affect the performance goals are highly correlated with states that are easy to measure (e.g., speed), then the more advanced planning methods will offer a large benefit, but if internal states are not correlated, there is less advantage. This observation motivates the need for detailed understanding of the actual distribution of behavior models and internal states in the driver population, especially correlation between features.

There are many possible extensions to this work. It may be beneficial to try more advanced POMDP solvers such as DESPOT [23]. Thus far, only inter-vehicle internal state heterogeneity has been considered, but real drivers also change their behavior over time and in response to stimuli [24], and it may be useful to model this when planning. Moreover, it may be even more beneficial to take a driver's specific intentions into account when planning [25].